%% file: main.tex
\documentclass[conference]{IEEEtran}

\usepackage[utf8]{inputenc}
\usepackage[T1]{fontenc}
\usepackage{cite}
\usepackage{float}
\usepackage{booktabs}
\usepackage{multirow}
\usepackage{graphicx}
\usepackage{amsmath}
\usepackage{placeins}
\usepackage[caption=false,font=footnotesize]{subfig}
\usepackage{epsfig}
\usepackage{listings}
\usepackage{xcolor}

\definecolor{codebg}{RGB}{248,248,248}
\definecolor{codekeyword}{RGB}{0,92,197}
\definecolor{codestring}{RGB}{163,21,21}
\definecolor{codecomment}{RGB}{0,128,0}
\definecolor{codenumber}{RGB}{120,120,120}

\title{A Knowledge-Injection Framework for Zero-Shot Adaptation of LLMs to Delirium Prediction}

\author{
\IEEEauthorblockN{Jessica Sena, Shesadree Priyadarshani, Miguel Contreras, Bharat Gandhi, \\Scott Siegel, Subhash Nerella, and Parisa Rashidi}
\IEEEauthorblockA{Herbert Wertheim College of Engineering, University of Florida\\
Gainesville, FL, USA}
}

\begin{document}

\maketitle

\begin{abstract}
Large language models show promise for clinical prediction, but zero-shot performance on specialized tasks is limited by incomplete domain knowledge, especially for smaller locally deployable models. We present a lightweight knowledge-injection framework for zero-shot ICU delirium prediction that augments a deterministic natural-language summary of structured electronic health record data with an external clinical knowledge report at inference time, without fine-tuning or retrieval. We evaluate LLaMA 3.1 8B and LLaMA 3.3 70B on 3,160 ICU admissions from the MIMIC IV dataset. Adding a clinically meaningful external knowledge report improves AUROC by 8.57 percentage points for the 8B model and 1.99 percentage points for the 70B model compared to no external knowledge. Relative to a GPT-5.2 frontier-model reference without external knowledge report (AUROC 68.86\%), knowledge injection reduces the performance gap from 15.66 to 7.09 AUROC points for LLaMA 8B and from 5.30 to 3.31 AUROC points for LLaMA 70B. Random control reports do not improve performance and often degrade it, indicating that gains depend on clinically meaningful content rather than added prompt length alone. SHAP-based attribution further confirms that the injected knowledge is actively used during prediction. These findings suggest that inference-time knowledge injection can narrow the gap between locally deployable open-weight models and frontier closed models while preserving a practical, privacy-preserving workflow for resource-constrained clinical settings.
\end{abstract}

\input{introduction.tex}
\input{method.tex}
\input{results.tex}
\input{discussion.tex}
\input{conclusions.tex}

\begingroup\sloppy
\bibliographystyle{IEEEtran}
\bibliography{references}
\endgroup

\appendices
\input{appendix.tex}

\end{document}

%% file: introduction.tex
\section{Introduction}
\label{sec:intro}

Large language models (LLMs) are neural network models designed to understand and generate natural language at scale~\cite{brown2020language}. These models have rapidly expanded from general-purpose language tools into healthcare, where they are being explored across a broad range of applications including medical question answering, clinical documentation, patient communication, and decision support~\cite{thirunavukarasu2023large,omiye2024large,LIN2025100868,maity2025llms}. Recent reviews~\cite{bedi2025testing,zhang2025revolutionizing} show that LLM evaluations in healthcare now span most medical specialties and a wide variety of tasks, underscoring the speed and breadth of clinical interest in these models. Within this broader landscape, clinical prediction is an especially important use case since it can directly support risk stratification and clinical decision-making. Prior work suggests that LLMs may be useful for this purpose: studies~\cite{jiang2023health, mccoy2024characterizing, ma2024hrbgcn, zhou2025frailty, shashikumar2025sepsis}  have shown that LLMs fine-tuned on clinical notes can be used for outcomes such as readmission, in-hospital mortality, and length of stay, while other works~\cite{amirahmadi2025toobert, groza2025noonan, xian2024patientembedding} suggest that they may also capture temporal and contextual signals from longitudinal records to forecast disease trajectories. Additional studies further indicate that prediction and phenotyping may be possible even in zero-shot settings, without task-specific training, which is particularly appealing in clinical environments where annotated data are scarce and prediction targets can shift over time~\cite{alsentzer2023zero,jin2024matching}. 

Despite this promise, pretrained general-purpose LLMs have well-documented limitations when applied to clinical prediction. They can hallucinate facts, show inconsistent accuracy, and lack the specialized knowledge needed for reliable inference, raising familiar concerns about accuracy, bias, interpretability, privacy, and regulation in real-world medical use~\cite{omiye2024large,briganti2024chatgpt,maity2025llms,zhang2025revolutionizing}. These limitations are especially pronounced in smaller LLMs, whose reduced parameter capacity restricts the amount of domain knowledge acquired during pretraining and, in turn, lowers performance on specialized clinical tasks~\cite{long2024chatent,temsah2025deepseek}. These limitations matter in practice because smaller models are often the only feasible option for on-premises deployment due to computational and budget constraints. At the same time, recent evidence suggests that compact open-source LLMs can approach the performance of much larger models when supplemented with external knowledge at inference time, making knowledge injection an attractive strategy for resource-efficient clinical deployment~\cite{bartels2025opensource}.

One common response is to fine-tune LLMs on task-specific clinical data. Although fine-tuning can improve predictive performance by aligning models with specialized data and objectives~\cite{jiang2023health,henriksson2023multimodal,ANISUZZAMAN2025100184}, it introduces important tradeoffs. Instruction tuning often changes response style more than underlying knowledge, and full-parameter fine-tuning can erode pretrained knowledge and even increase hallucinations~\cite{ghosh2024closerlooklimitationsinstruction}. In addition, fine-tuning requires substantial labeled data and compute, risks overfitting, and may reduce generalization across institutions or across changing clinical definitions~\cite{ANISUZZAMAN2025100184}. This dependence on annotated data is not unique to LLMs but a longstanding constraint shared by supervised machine learning and deep learning models in medicine more broadly, where high-quality labels typically require expert chart review or image annotation that is time-consuming, costly, and prone to inter-annotator variability~\cite{alzubaidi2021novel,pachetti2024systematic}. Even when sufficient labels are obtained, models trained at one institution often degrade on data from another because of dataset shift in patient populations, acquisition protocols, and documentation practices~\cite{wan2022unified,zhou2022multisite}. As Theodoris et al~\cite{theodoris2023transfer} note, such labeled data are especially scarce in rare diseases, emerging conditions, and sparsely sampled cohorts, further limiting the practicality of this approach.

External knowledge injection offer an alternative that does not require modifying model weights~\cite{wu2024llm}. The dominant paradigm in this area is retrieval-augmented generation (RAG), introduced by Lewis et al.~\cite{lewis2021retrieval}, which combines pretrained LLMs with document retrieval to ground outputs in external evidence and reduce hallucinations~\cite{DBLP:journals/corr/abs-2104-07567,yang2025retrieval}. In clinical settings, RAG has improved factuality and utility in domain-specific question answering~\cite{zakka2024almanac,long2024chatent,ge2024development}, and graph-based extensions further support relational reasoning in biomedical question answering and decision support~\cite{zhao2025medrag,wu2025medical}. A recent systematic review and meta-analysis by Liu et al.~\cite{liu2025improving} found that RAG improves LLM performance across a range of biomedical tasks. Still, RAG brings practical drawbacks: performance depends heavily on retrieval quality and corpus coverage, retrieved evidence can be noisy or fragmented, and the added retrieval pipeline increases latency and system complexity, which can make deployment less straightforward for structured clinical prediction.

Complementary to retrieval-based and graph-augmented approaches, knowledge-aware zero-shot learning seeks to improve generalization by explicitly incorporating structured external knowledge, such as ontologies and domain reports, rather than relying only on data-driven correlations~\cite{chen2021knowledgeawarezeroshotlearningsurvey}.Yet recent evidence suggests that gains seen in broader LLM applications do not translate cleanly to core biomedical tasks \cite{nagar-etal-2025-llms,omiye2024large}. While LLMs can approach expert-level performance on clinical question answering and summarization, they remain limited on tasks that require precise structured outputs, such as zero-shot biomedical information extraction, and advanced prompting or retrieval often yields little improvement~\cite{nagar-etal-2025-llms}. This limitation is especially important for structured clinical prediction, which has received far less attention than text-generation applications. Moreover, prior studies rarely test whether improvements come from the semantic relevance of the injected knowledge or simply from adding more text to the prompt through comparisons with random or irrelevant context~\cite{wen-etal-2024-characterizing}. It also remains unclear how these effects vary with model size, an issue with direct practical importance for smaller, resource-efficient clinical deployments.

In this work, we address these gaps by examining whether external clinical knowledge injection can improve prediction without requiring model retraining or complex retrieval pipelines. A central motivation is usability: clinicians who identify relevant prediction problems often do not have deep expertise in machine learning, whereas approaches such as fine-tuning or RAG typically require specialized technical knowledge, infrastructure, and maintenance. We therefore focus on a clinician-friendly framework in which task-relevant knowledge is supplied at prediction time directly to an off-the-shelf LLM model, rather than being incorporated through retraining, with the goal of preserving simplicity while improving task-specific reasoning.

We study this question in the context of delirium prediction in the ICU. Delirium is a common, multifactorial, and clinically relevant syndrome for which risk depends on heterogeneous physiological, laboratory, medication, and comorbidity information~\cite{wilson2020delirium}. This complexity makes delirium a demanding test case for zero-shot clinical inference. Using 3,160 ICU stays from the MIMIC IV dataset~\cite{johnson2023mimic}, we evaluate whether structured external knowledge can improve delirium prediction with off-the-shelf LLMs, whether any observed gains depend on the semantic relevance and format of the injected knowledge, and how much this approach can narrow the gap to a frontier closed-model reference. To the best of our knowledge, this is the first study to evaluate this form of inference-time external knowledge injection for zero-shot structured clinical prediction of delirium.

The main contributions of this work are summarized as follows:
\begin{enumerate}
    \item We introduce a lightweight inference-time knowledge-injection framework for zero-shot delirium prediction from structured EHR summaries that does not require model fine-tuning or a retrieval pipeline.
    \item We provide evidence that the benefit of inference-time external knowledge depends on both semantic relevance and model scale, showing that clinically meaningful knowledge is more useful than token-matched random context and that smaller models stand to benefit most.
    \item We show that knowledge injection also narrows the gap to a GPT-5.2 data-only frontier-model reference, reducing the AUROC gap from 15.66 to 7.09 points for LLaMA 8B and from 5.30 to 3.31 points for LLaMA 70B.
    \item We show that the structure of injected knowledge materially affects model behavior, and we use SHAP-based attribution analysis to provide insight into whether and how much the model incorporates the injected knowledge during prediction.
\end{enumerate}

%% file: method.tex
\section{Methods}
\label{sec:methods}

\begin{figure*}[bt]
    \centering
    \includegraphics[width=\textwidth]{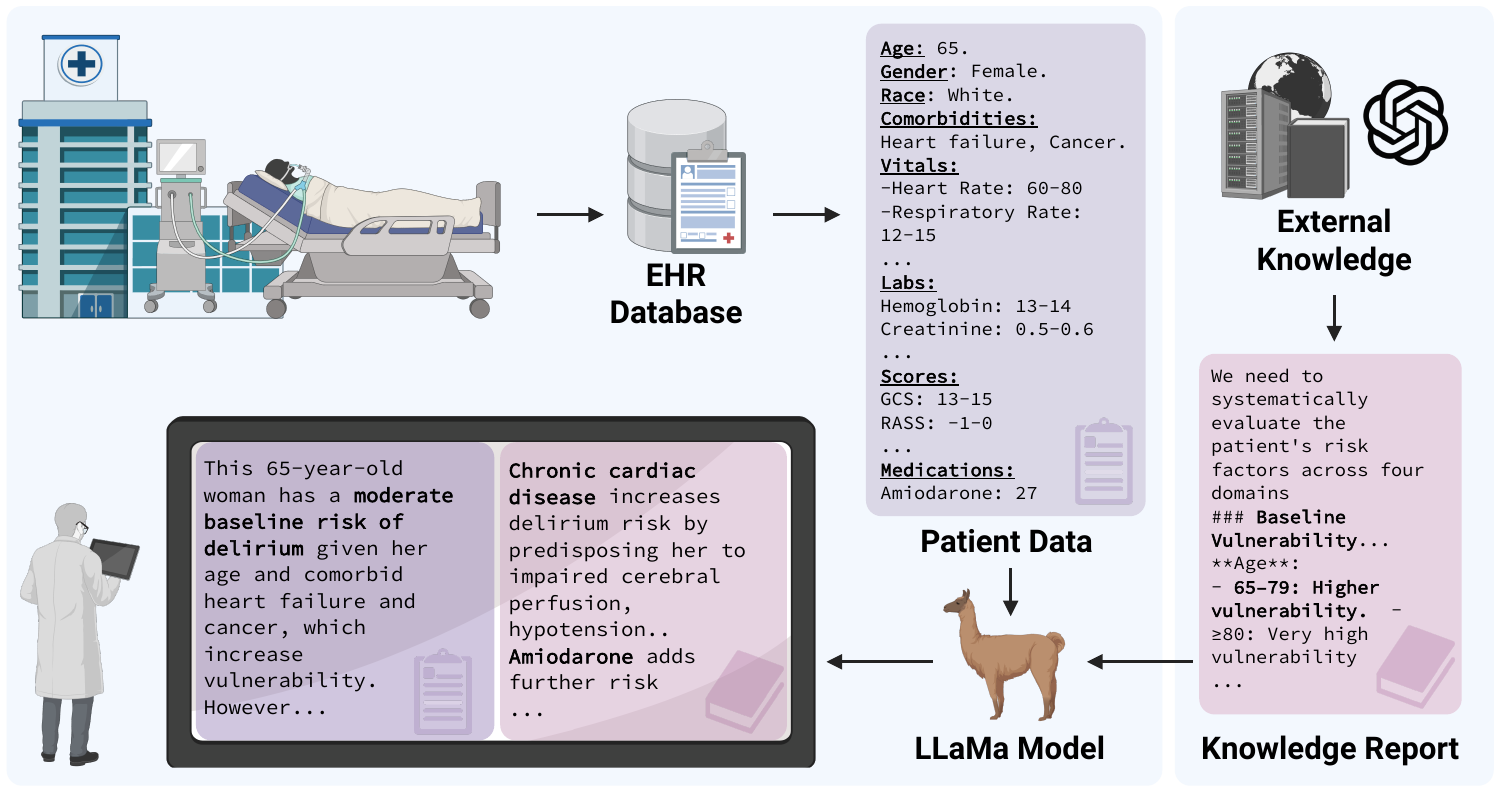}
    \caption{Clinical workflow for delirium risk assessment using an LLM. Patient-specific information from the EHR is summarized into a structured report and combined with a brief clinical knowledge report describing the task, for example delirium risk factors. Both are provided directly to an off-the-shelf LLM to generate patient-specific predictions. This approach enables clinicians to adapt general-purpose LLMs to clinical prediction tasks without requiring model fine-tuning, retrieval pipelines, or specialized computational expertise, simplifying deployment and use in real-world settings.}
    \label{fig:overview}
\end{figure*}

\subsection{Data and study design}

\begin{figure}
    \centering
    \includegraphics[width=1\linewidth]{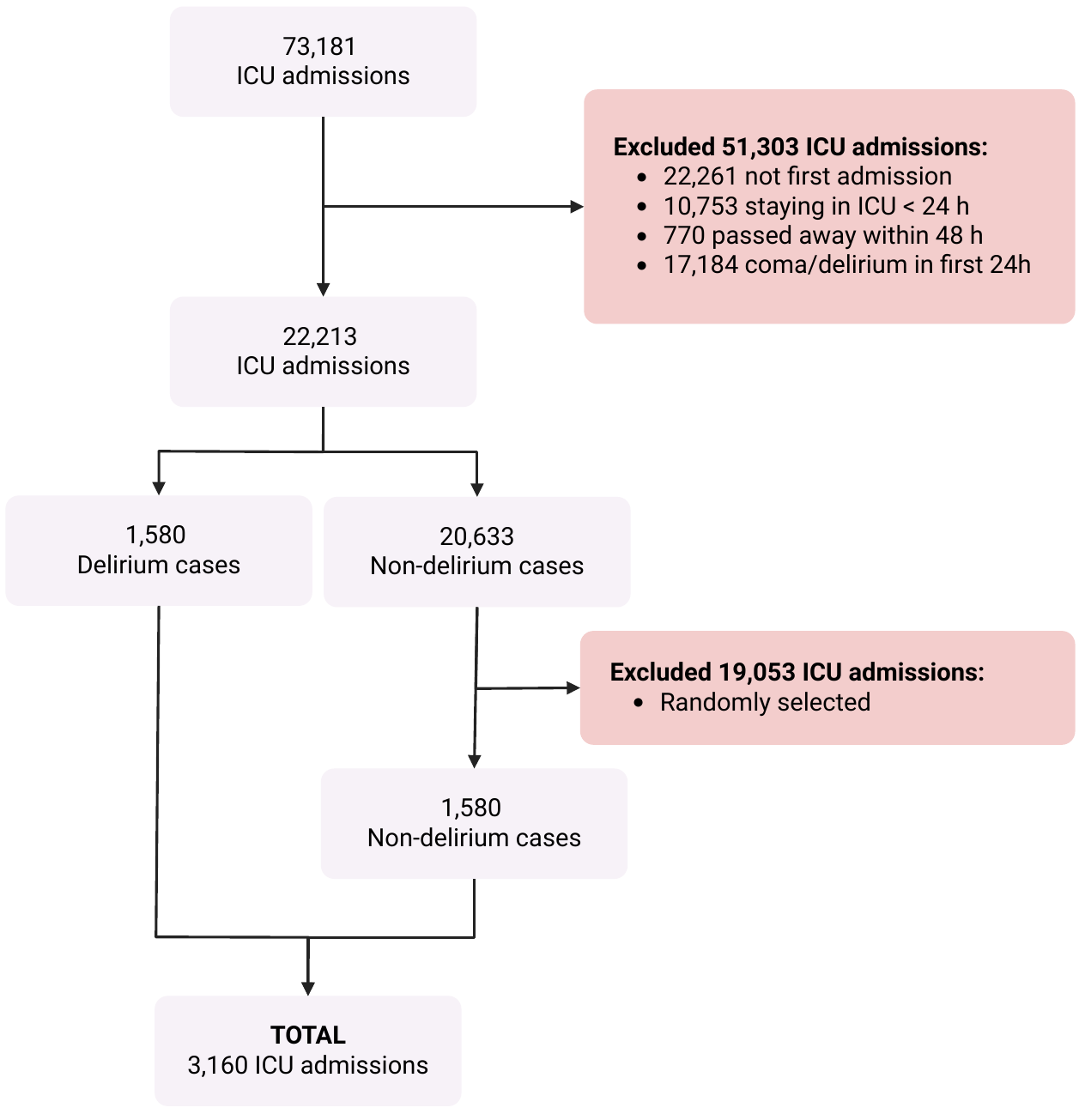}
    \caption{Inclusion and exclusion criteria for MIMIC IV dataset.}
    \label{fig:mimic}
\end{figure}

We conducted a retrospective analysis using data from the Beth 
Israel Deaconess Medical Center collected between 2008 and 2019, 
obtained from the publicly available Medical Information Mart for 
Intensive Care (MIMIC IV)~\cite{johnson2023mimic} dataset. ICU 
admissions were excluded if they were not the patient's first ICU 
stay or if the ICU length of stay was less than 24 hours. These 
criteria were applied to reduce bias associated with repeated ICU 
admissions and to ensure sufficient data availability. To limit 
bias related to high illness severity at admission, we excluded 
patients who died within 48 hours of ICU admission or who 
exhibited delirium or coma within the first 24 hours. ICU stays 
without available electronic health record (EHR) data during the 
initial 24-hour period were also excluded. A description of the 
cohort selection is shown in Figure~\ref{fig:mimic}.

After applying these exclusion criteria, the full cohort comprised 
22,213 ICU admissions. The dataset exhibited substantial class 
imbalance, with delirium-positive cases representing a minority 
of the cohort. To enable efficient zero-shot evaluation under 
computational time and cost constraints, a balanced evaluation set 
was constructed by randomly undersampling the majority class, 
yielding 3,160 ICU stays with equal numbers of delirium-positive 
and delirium-negative cases (1,580 each). All zero-shot model 
evaluations were conducted on this balanced set.

\subsection{Features and Outcomes}

The objective was to predict whether a patient would develop delirium at any time after the first 24 hours of ICU admission. This 24-hour window was used to establish a clinically meaningful baseline of physiologic, laboratory, and treatment-related information, enabling early risk stratification and potential proactive clinical management. Delirium was defined as the presence of a positive Confusion Assessment Method for the ICU (CAM-ICU) result in conjunction with a Richmond Agitation–Sedation Scale (RASS) score of $-3$ or higher~\cite{gusmao2012confusion} during any 12-hour interval occurring after the initial 24 hours of ICU stay.

Predictive features included both temporal and static patient data. Temporal variables were extracted from the first 24 hours of ICU admission and comprised four categories: vital signs, laboratory measurements, medication administrations, and clinical assessment scores. Static variables consisted of demographic characteristics and comorbidity information obtained from records available at admission time. A total of 81 predictive features were used for delirium prediction. This features were in categories such as demographics, comorbidities, assessment scores, laboratory values, medication and vital signs. For more details please refer to~\cite{contreras2025large}.

\subsection{Method Development and Performance}

The proposed framework formulates delirium prediction in the ICU as a zero-shot inference task. The LLMs methods employed in the evaluation are not fine-tuned on the target dataset; instead, task-specific information is provided at inference time through a combination of (i) an external knowledge report about delirium and (ii) a structured, text-based summary of each patient's electronic health record (EHR) data.

As illustrated in Figure~\ref{fig:overview}, our workflow consists of three main stages. First, an external knowledge report (see Section~\ref{subsec:external-report}) is generated once for the delirium prediction task by prompting a GPT~5.2~\cite{openai_chatgpt_2026} model with a description of the target outcome and the relevant clinical feature categories; this report is reused across all patient inferences. Second, for each ICU stay, a structured EHR summary (see Section~\ref{subsec:patient-data-report}) is constructed by aggregating time-varying clinical measurements, summarized by taking the minimum and maximum values for each feature during the first 24 hours of ICU admission, and static characteristics into a short narrative text. Third, the external knowledge report and the patient-specific EHR summary are combined into a single prompt (see Section~\ref{subsec:prompt}) and passed to an LLM. The prompt assigns the model the role of an ICU delirium risk assessment system and instructs it to identify which observed clinical findings increase or reduce delirium risk based on the provided knowledge report and patient data. The model is then asked to produce a structured chain-of-thought reasoning trace of up to three steps, followed by a probability score between 0 and 100 reflecting the likelihood that the patient will develop delirium before ICU discharge. This score is used as the basis for binary classification and evaluation. The full prompt and full external knowledge reports used in this work are available in Appendix.

To assess the effect of incorporating external knowledge across different model scales, we employed models from the LLaMA 3.x family of open-weight, decoder-only transformer architectures. Specifically, we used LLaMA 3.1 evaluating its 8B-parameter variant to provide a balance between representational capacity and computational efficiency. 
We also included LLaMA 3.3, the most recent release in the LLaMA 3.x family, which incorporates additional training and alignment refinements; its 70B-parameter variant was used to characterize the behavior of high-capacity models.

\subsection{Structured EHR Summary Generation}
\label{subsec:patient-data-report}
Following the work of Contreras et al.~\cite{contreras2025large}, we convert structured EHR data into text for use with LLMs and represented each ICU stay as a deterministic narrative summary rather than as raw tabular input. Time-varying variables from the first 24 hours are aggregated into concise textual descriptions, primarily using observed minimum and maximum values, while static demographic and comorbidity variables are rendered as short declarative statements. These components are concatenated into a standardized patient data report that preserves the main clinical content of the structured record while matching the natural-language input format expected by the LLM; Figure~\ref{fig:overview} illustrates this representation.

\subsection{External Knowledge Report}
\label{subsec:external-report}
\subsubsection{Real Report}

The external knowledge report is a task-level text that encodes general clinical knowledge about delirium in the ICU. The goal is to provide the model with task-specific knowledge without fine-tuning it on the target task. The report is generated prior to any patient-level inference and reused across all patients.

We generated two report variants using GPT-5.2 under different generation settings. The first variant (v1) was produced using GPT-5.2 under standard prompting, in which the model relies 
solely on its parametric knowledge to synthesise clinical content. The second variant (v2) was produced using GPT-5.2 with its deep research mode enabled, which augments the generation process with 
iterative retrieval from external sources, allowing the model to ground its output in recent literature and to expand coverage of 
relevant clinical factors. Both variants were produced using the same underlying model, isolating the effect of grounded, 
retrieval-augmented generation relative to parametric-only generation.

The two variants differ in both content and structure. The v1 report presents a structured probability computation framework organised around four evidence domains: baseline vulnerability, 
acute physiologic instability, neurocognitive state and arousal, 
and medication-related risk proxies alongside protective signals and a probability mapping scale. Risk factors within each domain are described at a qualitative and categorical level without explicit numerical thresholds, and the probability mapping uses broad categorical anchors with a starting baseline of 15\%. 
This provides a flexible reasoning scaffold that leaves the model room to integrate continuous clinical evidence without being 
constrained by specific cutoffs.

The v2 report is organised around the same four domains but introduces explicit numerical thresholds for nearly every risk 
factor. Examples include graded age bands (65--79 versus $\geq$80 years), specific vasopressor dose thresholds (norepinephrine-equivalent $>$0.1~$\mu$g/kg/min), organ dysfunction flags defined by concrete laboratory cutoffs 
(creatinine $\geq$2.0, bilirubin $\geq$2.0, platelets $<$100), and SOFA-style GCS severity bands (GCS 10--12 moderate, 6--9 
severe). The probability mapping is similarly more granular, with six defined risk tiers and multi-organ dysfunction counts as explicit anchors. The v2 report also incorporates additional clinical nuance drawn from retrieved literature, including the distinction between dexmedetomidine and propofol in terms of 
delirium risk, and the use of RASS $\leq -3$ as both a risk indicator and a limitation on protective inference.

The full text of both reports is provided in Appendix~\ref{appendix:report-v1} and 
Appendix~\ref{appendix:report-v2}.



\subsubsection{Random Report}
To isolate the effect of report presence from its semantic content, a synthetic \emph{random report} is constructed with a tokenized length identical to that of the corresponding real report. This approach ensures that both reports share the same length and structural constraints, differing exclusively in the information conveyed.

The random report is generated by uniformly sampling neutral placeholder words, similar to \textit{"lorem ipsum"}, that lack task-relevant meaning. These words are selected so that, when preceded by whitespace, each corresponds to a single tokenizer token, enabling precise control over the token count. Special sequence-boundary tokens are inserted as required by the tokenizer.

After generation, the text is first decoded and then tokenized again to capture any normalization effects of the tokenizer. It is then adjusted by either truncation or extension to exactly match the tokenized length of the real report. This process guarantees that random and real reports are perfectly aligned in length, allowing controlled comparisons where any performance differences can be attributed exclusively to semantic content rather than report length.

\subsection{Prompt Template for Zero-Shot Prediction}
\label{subsec:prompt}

For each ICU stay, we instantiate a standardized prompt that fuses 
two information sources: an external knowledge report and a structured summary of the patient's EHR. The template is held 
constant across all instances so that variation arises only from the ICU-stay specific content. In this formulation, the external report provides contextual guidance relevant to delirium risk, 
while the EHR summary supplies the patient-specific evidence that 
the model must weigh in light of that context.

The prompt assigns the model the role of an ICU delirium risk assessment system and instructs it to identify which observed 
clinical findings increase or reduce delirium risk based on the provided knowledge report and patient data. The external report text (Section~\ref{subsec:external-report}) and the structured EHR summary are inserted into the template in a fixed order, report first, patient summary second. The model is then asked to produce a structured chain-of-thought reasoning trace of up to 
three steps, each limited to 15--20 words, followed by a probability score between 0 and 100 reflecting the likelihood 
that the patient will develop delirium before ICU discharge. The output format is strictly enforced: the model is instructed 
not to output any code, pseudocode, markdown, headings, or explanations beyond the specified format.

For evaluation, the probability score
output by the model is used directly as a continuous prediction for AUROC computation. For threshold-based metrics (accuracy, precision, recall, and F1), 
a fixed threshold of 50 is applied, corresponding to equal weighting of positive and negative predictions on the 0--100 scale. This procedure does not involve fine-tuning, calibration, 
or any form of parameter update. The same underlying LLM is applied to every ICU stay, with the prompt content serving as 
the sole mechanism for task conditioning. The full prompt template is provided in Appendix~\ref{appendix:prompt-code}.

\subsection{Evaluation and Statistical Analysis}
\label{subsec:statistical-analysis}

Model prediction performance was evaluated primarily using the 
Area Under the Receiver Operating Characteristic Curve (AUROC). 
AUROC was selected as the primary metric for three reasons. 
First, the model outputs a continuous probability score on a 
0--100 scale rather than a hard binary prediction, and AUROC 
directly measures how well the model ranks positive cases above 
negative ones without requiring a threshold decision, avoiding 
the arbitrary choice of a cutoff that can inflate or deflate 
accuracy-based metrics depending on where it is set. Second, 
AUROC captures discrimination across all possible operating 
points rather than at a single fixed threshold, which is 
important in clinical settings where the optimal risk cutoff 
may vary across institutions or clinical contexts, and is 
robust to class imbalance in the evaluation cohort. Third, 
AUROC is a standard metric in clinical prediction research 
\cite{yildiz2025llms,jiang2023health,bedi2025testing}, 
facilitating interpretation of results in the context of the 
broader literature. To provide a complete characterisation of 
predictive performance, accuracy, precision, recall, and F1 
score were additionally computed at a fixed threshold of 50 
on the probability scale.

To quantify uncertainty in model performance and assess 
variability arising from the stochastic nature of LLM 
inference, a 20-iteration repeated evaluation procedure was 
performed for each experimental condition. In each iteration, 
the balanced evaluation cohort of 3,160 ICU stays was 
resampled with replacement using a fixed random seed tied to 
the iteration index, ensuring that all conditions within the 
same iteration were evaluated on identical patient samples. 
This design measures model output variance across different 
data realisations. The 95\% confidence interval (CI) for 
AUROC was derived from the 2.5th and 97.5th percentiles of 
the resulting distribution, and the mean across iterations 
was used as the point estimate for each condition. This 
procedure was applied consistently across all model scales, 
report versions, and report types.

To assess whether performance differences between each 
knowledge-augmented condition and the patient data only 
baseline were statistically significant, DeLong's test for 
correlated ROC curves \cite{delong1988} was applied within 
each bootstrap resample, where all conditions share identical 
patient samples, enabling a paired comparison of ROC curves. 
Results are summarised as the median $z$-statistic and median 
$p$-value across the 20 bootstrap resamples, along with the 
proportion of resamples in which $p < 0.05$. Significance 
was assessed at $p < 0.05$.

To assess whether differences in inference time across 
conditions were statistically significant, a two-sided 
Mann-Whitney U test was applied to the distribution of 
per-sample inference times across the 20 bootstrap 
repetitions, treating each run as an independent observation. 
Significance was assessed at $p < 0.05$.

To provide mechanistic insight into how the model utilised 
different sections of the input prompt, a SHAP-based 
attribution analysis was conducted on a subsample of 80 
patient samples. Mean absolute SHAP values per word were 
computed independently for the knowledge report and patient 
data sections, normalising for differences in section length 
to enable fair cross-section comparison. Within each report 
version, a Wilcoxon signed-rank test assessed whether 
attribution to the higher-contributing section was 
statistically significant. Between report versions, 
Mann--Whitney U tests were used to compare per-word 
attribution for the same section type across v1 and v2. 
Significance thresholds were set at $p < 0.01$ and 
$p < 0.001$. Together, these analyses were designed to 
assess not only whether knowledge injection improved 
performance, but also whether the injected knowledge was 
actively utilised by the model during prediction.

%% file: results.tex
\section{Results}
\label{sec:results}
We evaluated the proposed external knowledge injection framework on n=3,160 ICU admissions from the MIMIC clinical dataset in a zero-shot delirium risk prediction setting.

\subsection{Cohort characteristics}

\begin{table*}[t]
\centering
\setlength{\tabcolsep}{6.5pt}
\renewcommand{\arraystretch}{1.35}
\caption{Patient Characteristics}
\label{tab:pat_characteristics}
\begin{tabular}{@{}lrrrr@{}}
\toprule
& \multicolumn{1}{c}{Overall} 
& \multicolumn{1}{c}{No Delirium} 
& \multicolumn{1}{c}{Delirium} 
& \multicolumn{1}{c}{p-value} \\ 
\midrule
\textit{Basic Information} & & & & \\
Number of ICU admissions        & 3160 & 1580 & 1580 & N/A \\
Age, years, median (IQR)        & 67.0 (55.0--79.0) & 65.0 (52.0--76.0) & 70.0 (59.0--81.0) & $<$0.005 \\
Female, n (\%)                  & 1476 (46.7\%) & 743 (47.0\%) & 733 (46.4\%) & 0.75 \\
BMI, $kg/m^2$, median (IQR)     & 27.5 (23.7--31.9) & 27.8 (24.3--31.7) & 27.2 (23.4--31.9) & 0.12 \\
CCI, median (IQR)               & 0.0 (0.0--0.0) & 0.0 (0.0--0.0) & 0.0 (0.0--0.0) & 0.38 \\
ICU length of stay, days, median (IQR) & 3.3 (1.9--6.6) & 1.9 (1.4--3.1) & 5.9 (3.5--10.0) & $<$0.005 \\
\midrule
\textit{Race, n (\%)}           & & & & \\
African American                & 285 (9.0\%) & 140 (8.9\%) & 145 (9.2\%) & 0.80 \\
White                           & 2182 (69.1\%) & 1123 (71.1\%) & 1059 (67.0\%) & 0.02 \\
Other                           & 693 (21.9\%) & 317 (20.1\%) & 376 (23.8\%) & 0.01 \\
\midrule
\textit{Comorbidities, n (\%)}  & & & & \\
Liver disease                   & 110 (3.5\%) & 42 (2.7\%) & 68 (4.3\%) & 0.02 \\
Diabetes                        & 236 (7.5\%) & 110 (7.0\%) & 126 (8.0\%) & 0.31 \\
\begin{tabular}[c]{@{}l@{}}Chronic obstructive pulmonary \\ disease\end{tabular} 
                                & 247 (7.8\%) & 99 (6.3\%) & 148 (9.4\%) & $<$0.005 \\
Congestive heart failure        & 273 (8.6\%) & 127 (8.0\%) & 146 (9.2\%) & 0.25 \\
Cerebrovascular disease         & 140 (4.4\%) & 55 (3.5\%) & 85 (5.4\%) & 0.01 \\
Metastatic carcinoma            & 50 (1.6\%) & 27 (1.7\%) & 23 (1.5\%) & 0.67 \\
AIDS                            & 6 (0.2\%) & 1 (0.1\%) & 5 (0.3\%) & 0.22 \\
Renal disease                   & 204 (6.5\%) & 101 (6.4\%) & 103 (6.5\%) & 0.94 \\
\midrule
\textit{Outcomes, n (\%)}       & & & & \\
Coma                            & 614 (19.4\%) & 65 (4.1\%) & 549 (34.7\%) & $<$0.005 \\
Mortality                       & 353 (11.2\%) & 56 (3.5\%) & 297 (18.8\%) & $<$0.005 \\
\bottomrule
\end{tabular}
\vspace{4pt}

\parbox{0.98\textwidth}{\footnotesize 
\textit{Abbreviations:} N, number; BMI, body mass index; 
CCI, Charlson comorbidity index; IQR, interquartile range. 
\textit{P values:} P values for continuous variables are 
based on a two-sided Mann-Whitney U test. P values for 
categorical variables are based on Pearson's chi-squared test.}
\end{table*}

Cohort characteristics are summarized in 
Table~\ref{tab:pat_characteristics}. The balanced evaluation cohort comprised 3,160 ICU admissions (1,580 delirium-positive and 1,580 delirium-negative). Relative to patients who did not develop
delirium, those who developed
delirium were older (median age, 70.0 [IQR, 59.0--81.0] vs. 65.0 [52.0--76.0] years; $p<0.005$) and had a longer ICU length of stay (median, 
5.9 [IQR, 3.5--10.0] vs. 1.9 [1.4--3.1] days; $p<0.005$). Sex
distribution (46.4\% vs. 47.0\% female; $p=0.75$), BMI (27.2 vs. 27.8 $kg/m^2$; $p=0.12$), and Charlson comorbidity index were comparable between groups ($p=0.38$). Among 
comorbidities, liver disease (4.3\% vs. 2.7\%; $p=0.02$), chronic obstructive pulmonary disease (9.4\% vs. 6.3\%; $p<0.005$), and cerebrovascular disease (5.4\% vs. 3.5\%; $p=0.01$) were more prevalent in the delirium group, while 
diabetes, congestive heart failure, renal disease, metastatic carcinoma, and AIDS did not differ significantly between groups. Regarding race, the delirium cohort had a lower proportion of White patients (67.0\% vs. 71.1\%; $p=0.02$) and a higher proportion categorized as Other (23.8\% vs. 
20.1\%; $p=0.01$), while African American representation was similar (9.2\% vs. 8.9\%; $p=0.80$). Clinically, the delirium cohort had substantially worse outcomes, with 
higher rates of coma (34.7\% vs. 4.1\%) and in-hospital mortality (18.8\% vs. 3.5\%), both $p<0.005$.

\subsection{Model Performance}

\begin{table}[t]
\centering
\small
\setlength{\tabcolsep}{3pt}
\renewcommand{\arraystretch}{1.2}
\caption{Mean AUROC (\%) for delirium risk prediction with and 
without external knowledge.}
\label{tab:results}
\begin{tabular*}{\columnwidth}{@{\extracolsep{\fill}}lcrrrr@{}}
\toprule
\multicolumn{1}{c}{\begin{tabular}[c]{@{}c@{}}\textbf{External} \\ \textbf{Knowledge}\end{tabular}} & 
\multicolumn{1}{c}{\textbf{AUROC \%}} & 
\multicolumn{1}{c}{\textbf{95\% CI}} & 
\multicolumn{1}{c}{$\Delta$} &
\multicolumn{1}{c}{$p$} \\
\midrule
\multicolumn{5}{c}{\textbf{LLaMA 8B}} \\
\midrule
Patient Data Only & 53.20 & [51.50, 55.09] & --    & --       \\
Real (v1)         & \textbf{61.77} & [60.52, 62.91] & +8.57 & $<$0.001 \\
Random (v1)       & 49.83 & [48.69, 50.83] & -3.37 & 0.005    \\
Real (v2)         & 54.73 & [53.46, 56.02] & +1.53 & 0.163    \\
Random (v2)       & 52.64 & [51.31, 53.86] & -0.56 & 0.537    \\
\midrule
\multicolumn{5}{c}{\textbf{LLaMA 70B}} \\
\midrule
Patient Data Only & 63.56 & [62.01, 64.61] & --    & --       \\
Real (v1)         & \textbf{65.55} & [63.99, 66.89] & +1.99 & 0.023    \\
Random (v1)       & 63.01 & [61.63, 64.35] & -0.55 & 0.357    \\
Real (v2)         & 64.11 & [62.47, 65.55] & +0.55 & 0.561    \\
Random (v2)       & 63.08 & [61.45, 64.59] & -0.48 & 0.296    \\
\bottomrule
\end{tabular*}
\vspace{4pt}

\parbox{0.98\columnwidth}{\footnotesize
Values are means and 95\% confidence intervals estimated over 
20 bootstrap repetitions. $\Delta$ indicates change relative 
to the no-knowledge baseline. $p$-values are median DeLong's 
test values across 20 bootstrap resamples.}
\end{table}

The evaluation of the external knowledge injection framework 
is summarized in Table~\ref{tab:results}. In all scenarios, 
patient data was provided to the model as a structured EHR 
summary; the baseline condition included no external knowledge 
report, enabling direct measurement of the report's impact on 
prediction performance.

For LLaMA 8B, the patient data only baseline achieves an AUROC of 
53.20\% (CI: 51.50--55.09). Incorporating the v1 real 
knowledge report yields a substantial gain, reaching 61.77\% 
(CI: 60.52--62.91; $\Delta=+8.57\%$). The v2 real report 
produces only a marginal improvement of 54.73\% 
($\Delta=+1.53\%$). Among random reports, v1 degrades 
performance to 49.83\% ($\Delta=-3.37\%$), while v2 produces 
a small decrease to 52.64\% ($\Delta=-0.56\%$). Statistical 
significance of these differences is examined in 
Section~\ref{subsec:significance}.

For LLaMA 70B, the patient data only baseline is considerably 
stronger at 63.56\% (CI: 62.01--64.61), reflecting the 
larger model's greater capacity to extract predictive signal 
from patient data alone. The v1 real report yields a modest 
gain, reaching 65.55\% ($\Delta=+1.99\%$). The v2 real 
report produces a small improvement of 64.11\% 
($\Delta=+0.55\%$). Random reports show negligible impact, 
with v1 at 63.01\% ($\Delta=-0.55\%$) and v2 at 63.08\% 
($\Delta=-0.48\%$), indicating that the 70B model is largely 
robust to irrelevant context. Across both model scales, the 
v1 real report consistently produces the largest gains, and 
real reports consistently outperform random reports.

\begin{figure*}[t]
\centering
\subfloat[LLaMA 8B Performance]
{\includegraphics[width=0.48\linewidth]{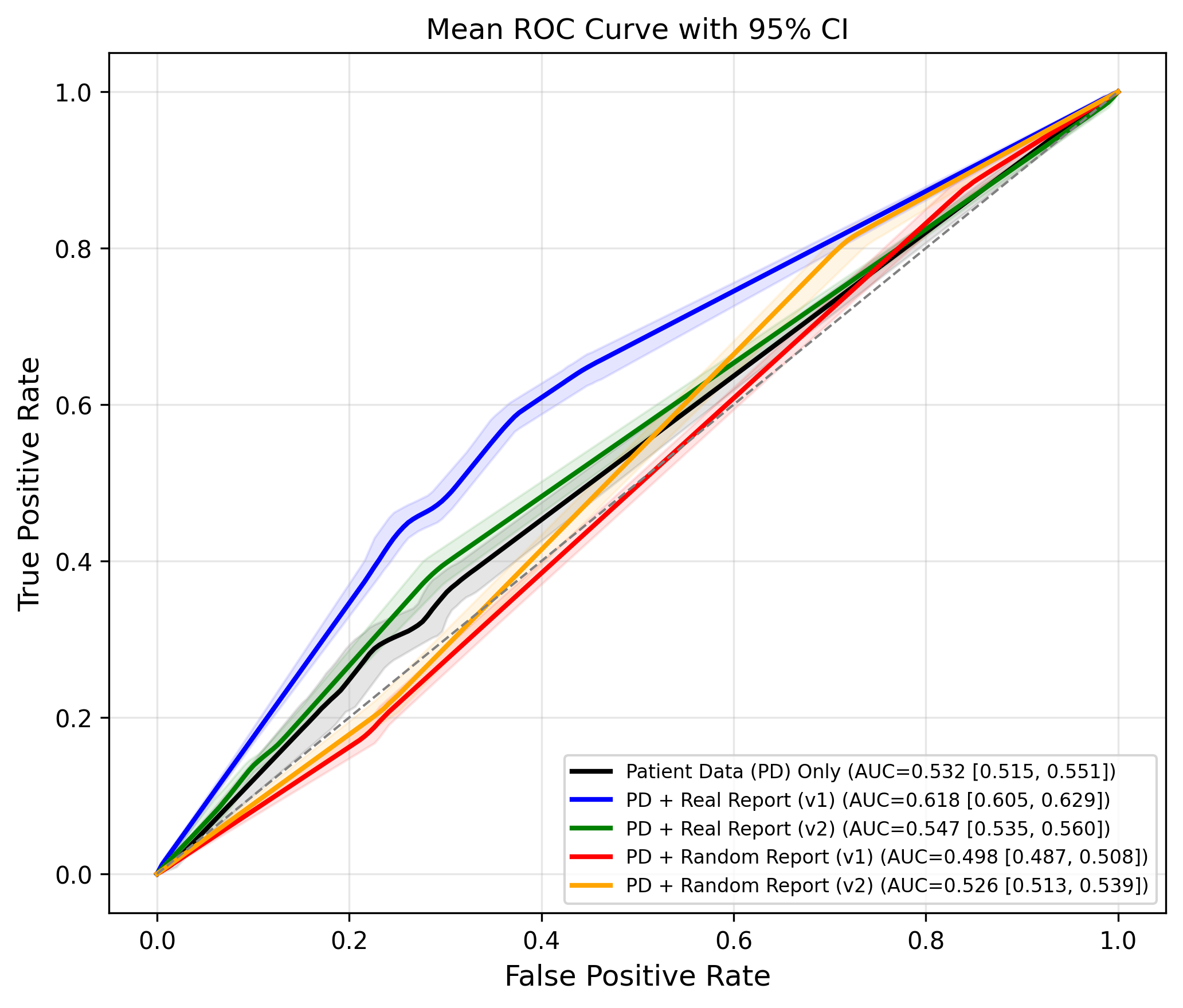}}%
\hfill
\subfloat[LLaMA 70B Performance]
{\includegraphics[width=0.48\linewidth]{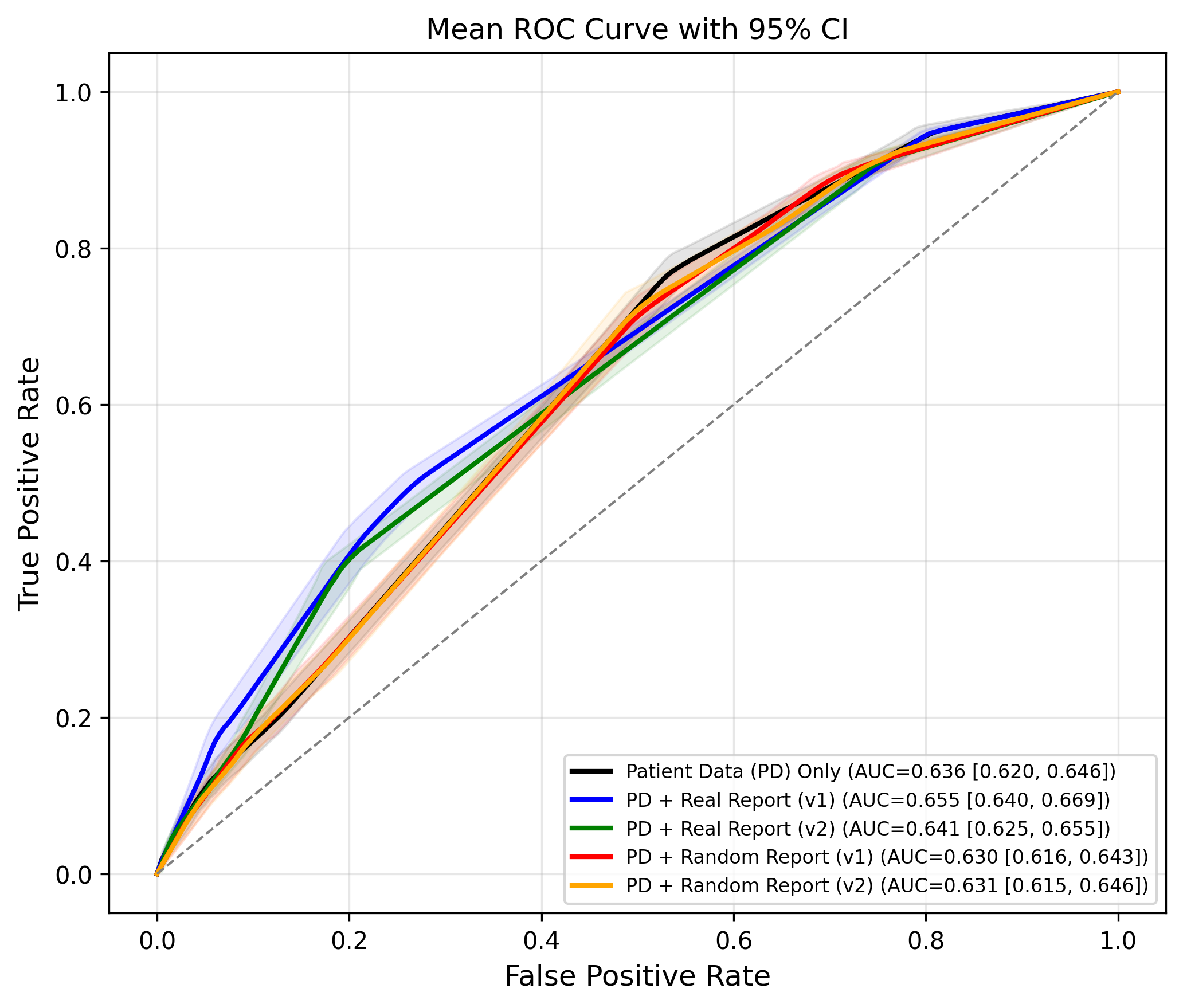}}
\caption{AUROC curve under different report conditions.}
\label{fig:model_performance}
\end{figure*}

\subsection{Statistical Significance of Performance Differences}
\label{subsec:significance}

Statistical significance was assessed using DeLong's test 
for correlated ROC curves \cite{delong1988}, applied within 
each bootstrap resample where all conditions share identical 
patient samples. Results are summarised as the median 
$p$-value across the 20 bootstrap resamples, along with 
the proportion of bootstraps in which $p < 0.05$.

For LLaMA 8B, the v1 real report is the only condition to 
achieve consistent and strong statistical significance 
(median $p<0.001$; prop$=1.00$), confirming that the 
observed gain is robust across all bootstrap resamples. 
The v1 random report shows significant degradation 
(median $p=0.005$; prop$=0.80$), indicating that 
irrelevant context harms the smaller model's 
performance. The v2 real report does not reach statistical 
significance (median $p=0.163$; prop$=0.35$), suggesting 
its modest improvement is not consistently detectable 
within individual resamples. For LLaMA 70B, the v1 real 
report reaches statistical significance (median $p=0.023$; prop$=0.70$), 
while no other condition achieves consistent significance, 
indicating that the 70B model is largely robust to both 
report type and content.

\subsection{Frontier Model Reference}

\begin{table}[h]
\centering
\small
\setlength{\tabcolsep}{4pt}
\caption{Performance of llama models with and without external knowledge compared to frontier commercial model}
\label{tab:frontier}
\begin{tabular}{llrr}
\toprule
\multicolumn{1}{c}{\textbf{Condition}} &
\multicolumn{1}{c}{\textbf{Model}} &
\multicolumn{1}{c}{\begin{tabular}[c]{@{}c@{}}\textbf{AUROC}\\\textbf{(\%)}\end{tabular}} &
\multicolumn{1}{c}{\begin{tabular}[c]{@{}c@{}}\textbf{Gap to }\\\textbf{GPT-5.2 (\%)}\end{tabular}} \\
\midrule
\multirow[c]{3}{*}{Patient Data (PD)}
& LLaMA 8B  & 53.20 & 15.66 \\
& LLaMA 70B & 63.56 & 5.30  \\
& GPT-5.2   & 68.86 & 0.00  \\
\midrule
\multirow[t]{2}{*}{%
\begin{tabular}[t]{@{}l@{}}
PD + External \\
knowledge report
\end{tabular}}
& LLaMA 8B  & 61.77 & 7.09 \\
& LLaMA 70B & 65.55 & 3.31 \\
\bottomrule
\end{tabular}
\vspace{4pt}

\parbox{0.98\columnwidth}{\footnotesize
Gap to GPT-5.2 is computed relative to the GPT-5.2 data-only AUROC of
68.86\%. GPT-5.2 is included as a proprietary frontier-model reference; its
parameter count and architecture have not been publicly disclosed. LLaMA
results are means over 20 bootstrap repetitions, whereas the GPT-5.2 result is
from a single full-cohort evaluation.}
\end{table}

To contextualize the effect of knowledge injection, we compared the
knowledge-augmented LLaMA models with a proprietary frontier-model reference
evaluated on the same balanced cohort. As shown in Table~\ref{tab:frontier}, GPT-5.2 achieved an AUROC of 68.86\% using as input only patient data. Although OpenAI has not publicly disclosed the parameter count or architecture of GPT-5.2, it represents an state-of-the-art model and therefore provides a useful upper-capability reference.

Knowledge injection substantially narrowed the gap between the smaller
open-weight models and this frontier-model reference. LLaMA 70B improved from
63.56\% with patient data alone to 65.55\% with the v1 real report, reducing
its absolute gap to GPT-5.2 from 5.30 to 3.31 AUROC points. The effect was
larger for LLaMA 8B: adding the v1 real report increased AUROC from 53.20\%
to 61.77\%, reducing the gap to GPT-5.2 from 15.66 to 7.09 AUROC points. Thus,
structured clinical knowledge allowed the 8B model to recover more than half
of its initial performance gap relative to the frontier-model reference,
despite its substantially smaller and explicitly bounded parameter scale.



\subsection{Inference Time}

\begin{table}[t]
\centering
\small
\setlength{\tabcolsep}{4pt}
\renewcommand{\arraystretch}{1.2}
\caption{Inference time across model sizes and report conditions}
\label{tab:inference_time}
\begin{tabular}{lrrr}
\toprule
\multicolumn{1}{c}{\textbf{Condition}} & \multicolumn{1}{c}{\begin{tabular}[c]{@{}c@{}}\textbf{Total}\\ \textbf{(h:m:s)}\end{tabular}} & \multicolumn{1}{c}{\textbf{95\% CI}} & \multicolumn{1}{c}{$p$} \\
\midrule
\multicolumn{4}{c}{\textbf{LLaMA 8B}} \\
\midrule
Patient Data Only  & 1:55:42 & [1:48:11, 2:28:33] & --       \\
Real (v1)          & 1:47:06 & [1:40:48, 2:14:54] & 0.007    \\
Real (v2)          & 1:42:19 & [1:38:04, 1:48:21] & $<$0.001 \\
Random (v1)        & 1:12:38 & [1:09:00, 1:21:42] & $<$0.001 \\
Random (v2)        & 1:14:32 & [1:10:09, 1:31:31] & $<$0.001 \\
\midrule
\multicolumn{4}{c}{\textbf{LLaMA 70B}} \\
\midrule
Patient Data Only  & 5:11:58 & [5:06:19, 5:19:10] & --       \\
Real (v1)          & 2:45:22 & [2:43:41, 2:47:09] & $<$0.001 \\
Real (v2)          & 2:51:27 & [2:49:11, 2:53:34] & $<$0.001 \\
Random (v1)        & 4:22:23 & [4:18:27, 4:25:40] & $<$0.001 \\
Random (v2)        & 4:32:18 & [4:29:37, 4:35:21] & $<$0.001 \\
\bottomrule
\end{tabular}
\vspace{4pt}

\parbox{0.98\columnwidth}{\footnotesize Values are means with 95\% CI over 
20 bootstrap repetitions. \mbox{$p$-values} from Wilcoxon signed-rank 
tests vs.\ data-only baseline.}
\end{table}

Inference time varies across report conditions and model scales, 
with statistically significant differences observed in all 
conditions relative to the patient data only baseline as reported in 
Table~\ref{tab:inference_time}. The 95\% confidence intervals 
for inference time were computed from the distribution of total 
inference times across the 20 bootstrap repetitions, using the 
2.5th and 97.5th percentiles of that distribution. For LLaMA 8B, 
total inference time ranges from 1:12:38 (CI: 1:09:00--1:21:42) 
with random reports to 1:55:42 (CI: 1:48:11--2:28:33) for the 
patient data only baseline, with real reports falling between at 1:47:06 
(v1) and 1:42:19 (v2). For LLaMA 70B, the data-only condition 
requires the longest inference at 5:11:58 (CI: 5:06:19--5:19:10), 
while real reports reduce total time substantially to 2:45:22 (v1) 
and 2:51:27 (v2), both significantly faster than baseline 
($p<0.001$). Random reports fall between these two extremes at 
4:22:23 (v1) and 4:32:18 (v2), also significantly faster than 
the patient data only condition ($p<0.001$).

\subsection{SHAP Section Contribution Analysis}

\begin{figure*}[t]
\centering
\includegraphics[width=\linewidth]{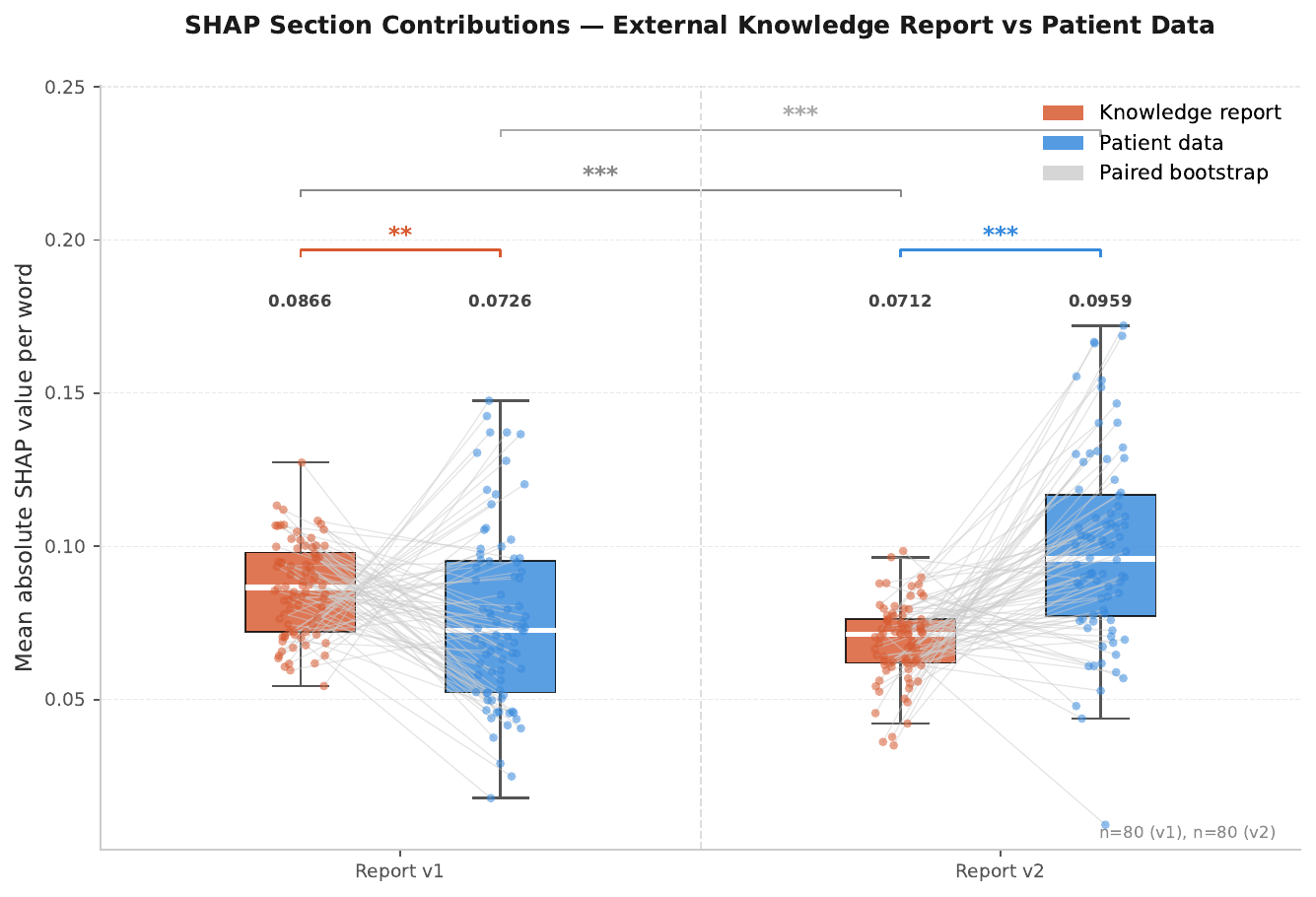}
\caption{Mean absolute SHAP value per word for the knowledge report 
and patient data sections across v1 and v2 report versions ($n = 80$ 
samples per version). Paired lines connect the same patient sample 
across sections. Brackets indicate significance levels from Wilcoxon 
signed-rank tests (within version) and Mann--Whitney U tests (between 
versions). {*}{*}{*}\,$p < 0.001$; {*}{*}\,$p < 0.01$.}
\label{fig:shap-boxplot}
\end{figure*}

To understand how the model distributes attention across input 
sections, we performed a SHAP-based attribution analysis on 80 
randomly selected patient samples, measuring the mean absolute SHAP value per word 
for the knowledge report and patient data sections independently. 
This metric quantifies per-token influence on the predicted delirium 
probability, controlling for differences in section length across samples.

For v1, the knowledge report contributes significantly more per word 
than the patient data (median 0.087 vs.\ 0.073; Wilcoxon 
$p = 7.03 \times 10^{-3}$). For v2, the pattern reverses: patient 
data contributes significantly more per word than the knowledge 
report (median 0.096 vs.\ 0.071; Wilcoxon $p = 1.99 \times 10^{-9}$). 
Cross-version comparisons show that the v1 knowledge report receives 
significantly higher per-word attribution than the v2 knowledge 
report (median 0.087 vs.\ 0.071; Mann--Whitney 
$p = 9.47 \times 10^{-11}$), while v2 patient data receives 
significantly higher per-word attribution than v1 patient data 
(median 0.096 vs.\ 0.073; Mann--Whitney $p = 1.61 \times 10^{-6}$).

%% file: discussion.tex
\section{Discussion}
\label{sec:discussion}

\subsection{Effect of Model Scale}

The results reveal a clear scale-dependent pattern in how LLMs respond 
to external knowledge injection. The 8B model begins from a weak 
baseline (53.20\% AUROC), which is close to random discrimination, 
suggesting that its parametric knowledge is insufficient to extract 
reliable delirium-relevant signal from structured patient data alone. 
The addition of a well-structured clinical knowledge report produces 
a substantial improvement (61.77\% AUROC), indicating that the model 
is able to leverage external guidance to compensate for what it cannot 
derive internally. The 70B model, in contrast, starts from a 
considerably stronger baseline (63.56\% AUROC), consistent with the 
view that larger models acquire more domain-relevant knowledge during 
pretraining and are better equipped to interpret clinical features 
without additional scaffolding 
\cite{temsah2025deepseek,long2024chatent}. As a result, the marginal 
benefit of knowledge injection is smaller (65.55\% AUROC), since much of the relevant signal is already captured by the model's parametric knowledge.


\subsection{Effect of Report Specificity}

Across both model scales, the concise v1 report consistently 
outperforms the more detailed v2 report, indicating that how 
knowledge is structured matters as much as how much is provided. 
Report v1 offers a flexible risk-assessment framework with qualitative 
categories and coarse adjustment ranges, leaving the model room to 
weigh clinical evidence across domains. Report v2, in contrast, 
provides granular numerical thresholds for nearly every risk factor 
and maps the final prediction to structured evidence-driver counts, 
which appears to constrain rather than guide the model's reasoning. 
This counterintuitive result is consistent with a well-documented 
limitation of LLMs: despite strong general language understanding, 
LLMs struggle with tasks that require precise numerical reasoning, 
threshold application, and rule-based quantitative decisions 
\cite{khandekar2024medcalc,mirzadeh2024gsmsymbolic,kim2025limitations}. 
In the clinical domain specifically, recent evidence shows that 
frontier models perform poorly on tasks requiring re-anchoring of 
numerical cutoffs to patient-specific context, relying instead on 
pattern-matched associations rather than principled threshold 
reasoning \cite{kim2025limitations,khandekar2024medcalc}. The 
granular thresholds in v2 specifying precise values such as 
MAP $<$70, FiO\textsubscript{2} $\geq$0.60, creatinine $\geq$2.0, 
and RASS $\leq -3$, may therefore have imposed a numerical 
reasoning burden the model could not reliably meet, forcing 
categorical decisions on features that are better treated as graded 
signals. This observation is further consistent with findings in 
cognitive science on human decision-making, where simpler and more 
targeted decision aids consistently outperform more comprehensive 
ones, and information beyond what is strictly needed for the 
decision at hand can distract and degrade performance 
\cite{kleinberg2023less}. The performance gap between v1 and v2 is 
more pronounced in the 8B model, which is consistent with evidence 
that smaller models exhibit greater sensitivity to prompt structure 
and formatting than their larger counterparts 
\cite{he2024promptformatting,sclar2024sensitivity}, suggesting that 
structural rigidity in external guidance disproportionately affects 
models with limited representational capacity. The 70B model exhibits 
greater but not unlimited robustness to report format, in line with 
findings that scaling reduces but does not eliminate prompt 
sensitivity \cite{sclar2024sensitivity}.

\subsection{Random versus Real Reports}

The comparison between real and random reports provides evidence that 
the observed gains reflect the semantic content of the knowledge 
report rather than any generic effect of adding text to the prompt. 
Random reports did not improve over the data-only baseline, and in 
the 8B setting they actively degraded performance, with the v1 
random report dropping to 49.83\% AUROC ($\Delta = -3.37\%$ relative 
to the 53.20\% baseline). This degradation suggests that smaller 
models do not simply benefit from the presence of auxiliary context they require that context to carry genuine task-relevant meaning. 
This is consistent with prior work showing that LLMs can be easily 
distracted by irrelevant context, which causes models to produce 
longer, less focused reasoning chains and degrades predictive 
performance \cite{shi2023distracted,wen-etal-2024-characterizing}. 
When presented with semantically unrelated text, the 8B model appears 
to allocate representational capacity toward processing irrelevant 
tokens, diluting rather than reinforcing the signal extracted from 
patient data. The 70B model is more robust to random reports, 
neither gaining substantially from real content (65.55\% with real 
v1 vs. 63.56\% baseline) nor suffering meaningful degradation from 
random content (63.01\% with random v1), consistent with the 
interpretation that larger models rely less on prompt context and 
more on internally encoded clinical knowledge 
\cite{temsah2025deepseek,yildiz2025llms}. Crucially, this controlled 
comparison rules out the possibility that performance gains are an 
artifact of increased prompt length or added context volume, and 
establishes that meaningful clinical knowledge is the active 
ingredient \cite{wen-etal-2024-characterizing}.

\subsection{Knowledge Injection as a Capacity-Compensation Strategy}
\label{sec:discussion_capacity_compensation}

The comparison with GPT-5.2 shows that structured clinical knowledge can move
smaller open-weight models substantially closer to frontier-model performance
in zero-shot delirium risk prediction. Although neither LLaMA model matched or
surpassed the GPT-5.2 data-only reference, the reduction in the performance gap
is notable. For LLaMA 8B, adding the v1 real report increased AUROC from
53.20\% to 61.77\%, reducing the gap to GPT-5.2 from 15.66 to 7.09 AUROC
points. LLaMA 70B also moved closer to the reference model, improving from
63.56\% to 65.55\% AUROC and reducing the gap from 5.30 to 3.31 points. Thus,
knowledge injection did not merely improve absolute performance; it also
changed the relative position of the open-weight models with respect to a
frontier-scale reference.

This result is important because frontier models likely encode substantially
more task-relevant clinical knowledge internally than smaller open-weight
models. Their larger training scale, broader exposure to medical and biomedical
content, and stronger general reasoning capacity may allow them to infer some
clinical relationships directly from patient data, even without an external
knowledge report. In contrast, smaller LLaMA models have more limited
parametric knowledge and weaker zero-shot reasoning capacity. The larger gain
observed for LLaMA 8B suggests that the external report supplied missing
clinical structure that the model could not reliably reconstruct on its own.
Rather than replacing model capacity, the injected knowledge appears to make the
available capacity more clinically useful.

The results therefore support knowledge injection as a capacity-efficient
strategy for zero-shot clinical prediction. In settings where frontier-scale
systems are difficult to deploy, structured clinical knowledge may help smaller
models recover part of the performance otherwise obtained through scale. This is
especially relevant for clinical environments, where smaller open-weight models
are generally cheaper to run and may be easier to host locally, audit, and
integrate within institutional data-governance constraints \cite{bartels2025opensource,wiest2025llmanonymizer}. 

\subsection{SHAP Attribution Analysis}

The SHAP attribution results provide a mechanistic account of the 
performance differences observed between report versions. For v1, 
the knowledge report receives higher per-word attribution than the 
patient data, indicating that the model actively draws on the 
clinical reasoning scaffold provided by the report when forming 
its prediction. For v2, the pattern reverses: the model shifts 
attribution weight away from the report and toward the patient data, 
suggesting that the more prescriptive structure of v2 is less 
effectively integrated. This shift in attribution is consistent 
with the AUROC results, where v2 yields a smaller performance gain 
than v1 when the report is not effectively utilised, the model 
falls back on patient data alone, approximating the data-only 
baseline. The cross-version comparisons further reinforce this 
interpretation: v1 reports attracted significantly higher per-token 
weight than v2 reports, while v2 patient data attracted 
significantly higher weight than v1 patient data.
These findings are consistent with prior work showing that 
semantically coherent and task-relevant context leads to more 
focused model reasoning \cite{struppek2025focusedcot}, and that 
irrelevant or poorly integrated context degrades predictive 
performance \cite{shi2023distracted}. The observed redistribution 
of attribution weight toward patient data when the knowledge report 
is less effectively integrated extends these findings by providing 
a mechanistic account of how models shift reliance across input 
sections when external guidance is not successfully incorporated a contribution enabled here by the token-level SHAP analysis. 
Taken together, these findings suggest that the performance 
advantage of v1 is not simply a consequence of report length or 
content volume, but reflects a qualitative difference in how 
effectively each report format is incorporated into the model's 
reasoning process.

\subsection{Inference Time}

An unexpected finding emerges from the inference time results: 
for the 70B model, real report conditions are substantially faster 
than the data-only baseline (2:45:22 and 2:51:27 for real v1 and 
v2, versus 5:11:58 for data only), while random reports fall 
between these two extremes (4:22:23 and 4:32:18). This pattern 
is counterintuitive, since longer prompts would normally be 
expected to increase inference time. One possible explanation 
lies in the generation mechanics of LLMs: when the input context 
provides structured, semantically coherent guidance, models tend 
to generate shorter and more focused reasoning chains before 
committing to a response, consistent with evidence that structured 
input context reduces token generation without sacrificing accuracy 
\cite{struppek2025focusedcot}. Ambiguous or unstructured context, 
by contrast, may introduce processing overhead that extends 
generation, though the precise mechanism behind this effect in 
our setting remains an open question. For the 8B model, inference 
time differences across conditions are smaller and less systematic, 
suggesting that this relationship between prompt content and 
generation efficiency is more pronounced at larger model scales. 
These timing results should be treated as exploratory and warrant 
further investigation under more controlled experimental 
conditions. Nevertheless, they suggest that knowledge-augmented 
prompting does not impose a computational overhead penalty in 
this setting and may in fact improve throughput for larger models.

\subsection{Implications for Clinical Practice}

The findings of this study suggest that lightweight knowledge injection 
at inference time may offer a practical pathway for supporting structured 
clinical prediction without requiring model retraining or retrieval 
infrastructure. In settings where fine-tuning is constrained by limited 
labeled data, computational resources, or shifting clinical definitions, 
providing a concise task-relevant knowledge report alongside patient data 
represents a low-overhead alternative that does not modify model weights 
\cite{ANISUZZAMAN2025100184,liu2025improving}. Our results demonstrate 
this effect in a specific and controlled setting such as zero-shot ICU delirium 
prediction using LLaMA models evaluated on MIMIC data and the 
degree to which these findings extend to other tasks or institutions 
remains to be established.

A distinctive advantage of this framework is that it is built entirely 
around locally deployable open-weight models, which has direct relevance 
for clinical settings where patient data privacy is a primary concern. 
Because inference runs on-premises, no patient data is transmitted to 
external servers or third-party APIs, a critical requirement in 
healthcare environments subject to regulatory constraints such as HIPAA 
and GDPR \cite{wiest2025llmanonymizer}. The only component that 
originates externally is the knowledge report, which contains no 
patient-specific information and is generated once and reused across 
all inferences. This stands in contrast to cloud-based commercial LLMs 
such as GPT, which require patient data to leave the local environment 
and introduce dependencies on external infrastructure, access policies, 
and model versioning that are difficult to control in clinical deployment 
\cite{zhang2025revolutionizing}. The combination of local execution, 
zero patient data exposure, and inference-time knowledge injection 
positions this framework as a privacy-preserving alternative that does 
not sacrifice the benefit of external domain knowledge.

The scale-dependent nature of the observed gains has further practical 
relevance. The 8B model showed a larger absolute improvement from 
knowledge augmentation compared to the 70B model, suggesting that 
smaller, resource-efficient models, which are more amenable to 
on-premises deployment given typical hospital computational budgets 
may benefit most from structured external knowledge at inference time. 
This is consistent with prior work showing that compact open-source 
models can narrow the gap with larger systems when supplemented with 
domain-relevant context \cite{bartels2025opensource,long2024chatent}. 
However, even with augmentation, the absolute performance levels 
observed here indicate that knowledge injection alone is unlikely to 
be sufficient for high-stakes clinical deployment without further 
validation and, potentially, complementary approaches.

The finding that the concise v1 report outperformed the more detailed 
v2 report, supported by SHAP attribution analysis, suggests that how 
knowledge is structured matters as much as how much is provided. 
This has a modest practical implication: clinicians contributing 
domain knowledge to such a framework need not produce exhaustive 
clinical guidelines, and a focused summary of key risk factors may 
be more effective. That said, the generalizability of this observation 
to other tasks, knowledge formats, or model families has not been 
tested here \cite{alsentzer2023zero,liu2025improving}.

\subsection{Limitations and Risks}

Several considerations should be noted when interpreting these results. 
The evaluation is conducted on a single retrospective cohort from one 
institution, and performance across different hospitals, EHR systems, 
or patient populations has yet to be established \cite{yildiz2025llms}. 
The quality and coverage of the external knowledge report are inherently 
dependent on the source used to generate it, and reports produced under 
different prompting strategies or knowledge bases may yield different 
results, an aspect that the v1 versus v2 comparison begins to 
explore but does not fully resolve. As a task-level construct, the 
knowledge report provides general clinical context that may not capture 
the full complexity of individual patient presentations, and this 
trade-off between simplicity and personalization is worth acknowledging. 
LLM behaviour can vary with prompt formulation, and while our controlled 
comparison with random reports provides evidence that semantic content 
drives the observed gains, systematic evaluation of prompt sensitivity 
remains an open area. Finally, as with any LLM-based clinical tool, 
deployment in real-world settings would require careful attention to 
interpretability, human oversight, and alignment with institutional 
and regulatory standards, considerations that are well recognized 
in the broader literature on clinical AI 
\cite{omiye2024large,maity2025llms,zhang2025revolutionizing}.

%% file: conclusions.tex
\section{Conclusion and Future Work}
\label{sec:conclusion}

We presented a zero-shot knowledge-injection framework for ICU delirium prediction that augments structured patient summaries with an external clinical knowledge report at inference time. In a held-out cohort of 3,160 ICU admissions from MIMIC, clinically meaningful external knowledge improved predictive performance relative to patient-data-only prompting, with the largest gains observed for the smaller LLaMA 8B model and more modest gains for the 70B model. Real knowledge reports consistently outperformed random reports, and the concise v1 report outperformed the more prescriptive v2 report, indicating that the usefulness of injected knowledge depends not only on its presence but also on how it is structured. Together, these findings suggest that inference-time knowledge injection is a lightweight and potentially practical approach for improving smaller open-weight LLMs on clinical prediction tasks without task-specific retraining.

Future work should evaluate this framework across external institutions, more diverse ICU populations, and additional clinical prediction tasks to establish generalizability. Further investigation is also needed to optimize the content and structure of injected knowledge, assess calibration and prompt robustness, compare this approach with retrieval-augmented and fine-tuned alternatives, and study prospective deployment with appropriate interpretability, clinician oversight, and workflow integration.

%% file: appendix.tex
\section{Report 1}
\label{appendix:reports}

\subsection{Report v1: Factor-Based Risk Framework}
\label{appendix:report-v1}

\begin{quote}
\small
\raggedright
\noindent To compute the probability of ICU delirium based on the 
provided evidence types, we need to systematically evaluate the 
risk factors across the domains you've outlined. Below is a 
step-by-step approach to adjust the probability based on the 
evidence provided:

\medskip
\noindent\textbf{Step 1: Baseline Vulnerability}
\begin{itemize}[noitemsep]
  \item Higher age: If the patient is elderly (e.g., $>$65 years), increase risk.
  \item Dementia or cerebrovascular disease: If present, increase risk.
  \item Higher Charlson index: If comorbidities are significant, increase risk.
  \item Systemic vulnerability (COPD, CHF, renal disease, liver disease, 
  metastatic cancer): If one or more of these conditions are present, increase risk.
\end{itemize}
\textit{Adjustment}: Mild baseline vulnerability: $+5$--$10\%$; 
Moderate: $+10$--$20\%$; Severe: $+20$--$30\%$.

\medskip
\noindent\textbf{Step 2: Acute Physiologic Instability}
\begin{itemize}[noitemsep]
  \item Hemodynamic instability: Low SBP/DBP, tachycardia, or vasopressors increase risk.
  \item Respiratory failure/high support: High FiO\textsubscript{2}, PEEP, or 
  abnormal EtCO\textsubscript{2}/SpO\textsubscript{2} increase risk.
  \item Metabolic stress: Elevated lactate, acidosis/alkalosis, or electrolyte 
  derangements increase risk.
  \item Inflammation/infection surrogate: Elevated WBC or CRP increases risk.
  \item Organ dysfunction: Elevated creatinine, bilirubin/AST/ALT, low albumin, 
  or abnormal INR/platelets increase risk.
\end{itemize}
\textit{Adjustment}: Mild instability: $+5$--$10\%$; 
Moderate: $+10$--$20\%$; Severe: $+20$--$40\%$.

\medskip
\noindent\textbf{Step 3: Neurocognitive State and Arousal}
\begin{itemize}[noitemsep]
  \item Lower GCS: If GCS is $<$13, increase risk.
  \item Agitation or fluctuating arousal (RASS high positive): If present, increase risk.
  \item Sedation exposure/deep sedation (RASS very negative + sedatives): 
  If present, increase risk.
\end{itemize}
\textit{Adjustment}: Mild neurocognitive risk: $+5$--$10\%$; 
Moderate: $+10$--$20\%$; Severe: $+20$--$40\%$.

\medskip
\noindent\textbf{Step 4: Medication-Related Delirium Risk Proxies}
\begin{itemize}[noitemsep]
  \item Sedatives/analgesics: Propofol, fentanyl, dexmedetomidine increase risk.
  \item High vasopressor burden: Any pressors or multiple pressors increase risk.
\end{itemize}
\textit{Adjustment}: Mild medication-related risk: $+5$--$10\%$; 
Moderate: $+10$--$20\%$; Severe: $+20$--$40\%$.

\medskip
\noindent\textbf{Step 5: Protective Signals}
\begin{itemize}[noitemsep]
  \item Stable vitals with narrow min--max ranges: Decrease risk.
  \item Minimal respiratory support: Decrease risk.
  \item No vasopressors: Decrease risk.
  \item Labs largely within normal/near-normal ranges: Decrease risk.
  \item Awake/calm arousal (RASS near 0) and higher GCS: Decrease risk.
\end{itemize}
\textit{Adjustment}: Mild protective signals: $-5$--$10\%$; 
Moderate: $-10$--$20\%$; Strong: $-20$--$30\%$.\\
\textit{Note: Protective signals should not be over-credited if sedation 
limits assessment reliability.}

\medskip
\noindent\textbf{Step 6: Combine Adjustments}\\
Start at \textbf{15\%} and adjust based on the evidence provided. 
Use the following ranges to guide the final probability:
\begin{itemize}[noitemsep]
  \item $5$--$15\%$: Stable physiology, minimal support, good arousal, low comorbidity.
  \item $20$--$40\%$: Moderate instability or vulnerability.
  \item $40$--$70\%$: Clear multi-domain instability and/or strong neurocognitive risk.
  \item $70$--$90\%$: Severe multi-organ instability, profound derangements, 
  deep sedation, very high vulnerability.
  \item $90$--$99\%$: Overwhelming evidence across all domains.
\end{itemize}

\medskip
\noindent\textbf{Example Calculation}\\
Patient profile:
\begin{itemize}[noitemsep]
  \item Baseline vulnerability: Elderly ($>$75 years), history of CHF and 
  renal disease $\rightarrow$ Moderate vulnerability ($+15\%$).
  \item Acute physiologic instability: On norepinephrine, FiO\textsubscript{2} 
  60\%, elevated lactate, creatinine 2.5 $\rightarrow$ Severe instability ($+30\%$).
  \item Neurocognitive state: GCS 10, RASS $-4$ (deep sedation) $\rightarrow$ 
  Severe neurocognitive risk ($+30\%$).
  \item Medication-related risk: On propofol and norepinephrine $\rightarrow$ 
  Moderate medication risk ($+15\%$).
  \item Protective signals: None strongly supported $\rightarrow$ No adjustment.
\end{itemize}
Starting probability: 15\%\\
Adjustments: $+15\%$ (baseline) $+30\%$ (instability) $+30\%$ 
(neurocognitive) $+15\%$ (medication)\\
\textbf{Final probability: 90\%} (severe risk of ICU delirium).
\end{quote}

\section{Report 2}
\subsection{Report v2: Feature-Conditioned Risk Framework}
\label{appendix:report-v2}

\begin{quote}
\small
\raggedright
\noindent To compute the probability of ICU delirium based on the 
provided evidence types, we need to systematically evaluate the 
patient's risk factors across four domains: baseline vulnerability, 
acute physiologic instability, neurocognitive state/arousal, and 
medication-related risk proxies. Protective signals can also be 
considered if strongly supported and assessable. Below is a 
step-by-step framework for applying this model:

\medskip
\noindent\textbf{Step 1: Baseline Vulnerability}
\begin{itemize}[noitemsep]
  \item Age: 65--79: Higher vulnerability; $\geq$80: Very high vulnerability.
  \item Dementia or baseline cognitive impairment/cerebrovascular disease: 
  Strong risk factor.
  \item Charlson comorbidity index: 1--2: Mild risk; 3--4: Moderate risk; 
  $\geq$5: Severe risk.
  \item Baseline organ vulnerability (liver disease, CKD/renal disease, COPD, 
  metastatic cancer): Moderate weight unless dementia is present.
\end{itemize}

\medskip
\noindent\textbf{Step 2: Acute Physiologic Instability}
\begin{itemize}[noitemsep]
  \item Cardiovascular dysfunction: MAP $<$70 or vasopressor use; high support: 
  norepinephrine-equivalent $>$0.1~$\mu$g/kg/min or $\geq$2 vasopressors.
  \item Shock physiology: vasopressors + lactate $>$2~mmol/L.
  \item Respiratory failure: mechanical ventilation (strong risk factor); 
  high oxygenation requirement (FiO\textsubscript{2} $\geq$0.60).
  \item Metabolic acidosis: pH $<$7.35 with low bicarbonate/base; 
  profound: pH $<$7.20.
  \item Organ dysfunction flags: creatinine $\geq$2.0, bilirubin $\geq$2.0, 
  platelets $<$100. Count the number of organ systems meeting these thresholds.
  \item Infection: Suspected/confirmed infection (strong risk factor).
\end{itemize}

\medskip
\noindent\textbf{Step 3: Neurocognitive State and Arousal}
\begin{itemize}[noitemsep]
  \item Low GCS: 10--12: Moderate risk; 6--9: Severe risk.
  \item Agitation or fluctuating arousal: RASS $\geq +2$ or marked fluctuation.
  \item Deep sedation/unassessable state: RASS $\leq -3$ (high risk).
\end{itemize}

\medskip
\noindent\textbf{Step 4: Medication-Related Delirium Risk Proxies}
\begin{itemize}[noitemsep]
  \item Benzodiazepines: Highest weight modifiable exposure.
  \item Continuous sedative infusion/deep sedation: RASS $\leq -3$ 
  regardless of agent.
  \item Vasopressor therapy: especially high-dose or multiple agents.
  \item Opioid infusion: moderate risk proxy.
\end{itemize}

\medskip
\noindent\textbf{Step 5: Protective Signals}\\
Consider protective factors only if strongly supported and assessable:
\begin{itemize}[noitemsep]
  \item Calm/awake arousal: RASS $-1$ to $+1$ with higher GCS.
  \item No vasopressors and no shock physiology.
  \item Minimal respiratory support and stable oxygenation needs.
  \item No metabolic acidosis and no moderate-to-severe organ dysfunction.
\end{itemize}

\medskip
\noindent\textbf{Step 6: Probability Mapping}\\
Start at a baseline probability of \textbf{15--20\%} and adjust based 
on the evidence:
\begin{itemize}[noitemsep]
  \item Low risk ($5$--$20\%$): Stable physiology, minimal support, 
  assessable calm arousal, low vulnerability.
  \item Moderate risk ($>20$--$40\%$): One major driver (e.g., 
  age/comorbidity, infection, mild-to-moderate organ dysfunction) 
  without deep sedation/coma.
  \item High risk ($>40$--$60\%$): $\geq$2 strong/moderate evidence 
  drivers (e.g., infection + sedation exposure; or mechanical 
  ventilation + another driver).
  \item Very high risk ($>60$--$80\%$): Mechanical ventilation and/or 
  deep sedation/coma plus additional drivers (shock physiology, 
  multi-organ dysfunction count $\geq$2).
  \item Extreme risk ($>80$--$95\%$): Ventilated + deep sedation/coma 
  + high cardiovascular support and/or multi-organ dysfunction 
  count $\geq$3.
  \item Overwhelming risk ($>95$--$99\%$): Only if evidence is 
  overwhelming and consistent across all domains.
\end{itemize}

\medskip
\noindent\textbf{Example Calculation}\\
\textit{Patient Data:}
\begin{itemize}[noitemsep]
  \item Baseline vulnerability: Age 78, Charlson score 4 (moderate), CKD present.
  \item Acute physiologic instability: MAP 65 with norepinephrine 
  0.15~$\mu$g/kg/min, lactate 3~mmol/L, FiO\textsubscript{2} 0.70, 
  creatinine 2.5, platelets 90 (multi-organ dysfunction count = 2).
  \item Neurocognitive state: GCS 8, RASS $-4$ (deep sedation).
  \item Medication-related risk proxies: Benzodiazepine infusion, 
  high-dose norepinephrine, opioid infusion.
  \item Protective signals: None (deep sedation limits assessment reliability).
\end{itemize}
\textit{Risk Assessment:}
\begin{itemize}[noitemsep]
  \item Baseline vulnerability: Moderate risk (age 78, Charlson 4, CKD).
  \item Acute physiologic instability: High risk (shock physiology, 
  multi-organ dysfunction count = 2, mechanical ventilation, 
  FiO\textsubscript{2} 0.70).
  \item Neurocognitive state: Severe risk (GCS 8, RASS $-4$).
  \item Medication-related risk proxies: High risk (benzodiazepines, 
  high-dose norepinephrine, opioids).
  \item Protective signals: None (deep sedation limits assessment).
\end{itemize}
\textit{Probability Mapping:}\\
Start at 15--20\%. Adjust for baseline vulnerability: $\sim$30\%. 
Adjust for acute physiologic instability: $\sim$50\%. 
Adjust for neurocognitive state: $\sim$70\%. 
Adjust for medication-related risk proxies: $\sim$80\%. 
No protective signals to lower risk.\\
\textbf{Final Probability: $\sim$80--85\% (very high risk).}
\end{quote}

\section{LLM Prompt code}
\label{appendix:prompt-code}

\begin{lstlisting}[language=Python]
def format_prompt(patient_summary):

    role = """
    Role: 
    You are an ICU delirium risk assessment system.
    """.strip()

    task = """
        Task:
        Using patient clinical data from the previous 24 hours and external clinical knowledge report if provided, identify which observed clinical findings indicate increased risk and which indicate reduced risk of INCIDENT delirium, and then estimate the probability (0-100) that the patient will develop delirium at any time before ICU discharge.
        """.strip()

    if report_text:
        knowledge_block = f"""
        External Clinical Knowledge Report:
        {report_text}
        """.strip()
    else:
        knowledge_block = ""

    knowledge_instructions = """
        Instructions:
        If an External Clinical Knowledge Report is provided above then base your prediction ONLY on risk or protective factors described in the report that are supported by the patient data. Do not assume or infer any missing patient attributes if not provided here.
        """.strip()

    patient_block = f"""
        Patient Clinical Data:
        {patient_summary}
        """.strip()

    patient_instructions = """
        Instructions:
        - Use only the observed patient data from the previous 24 hours.
        - The patient did NOT exhibit delirium during the previous 24 hours.
        - Some clinical features may be reported as min-max ranges, representing the lowest and highest observed values of that feature during this period.
        - Do not output any code, pseudocode, markdown, headings, or explanations.
        """.strip()

    output_format = """
        Respond in EXACTLY this format Strictly:
        **FORMAT INSTRUCTION: The output MUST STRICTLY follow the template below.**

        ### Chain of Thought:
        [Please sort out your thinking process step by step, with each logical step being a single sentence limiting strictly to 15-20 words only,and use a format such as <step 1> to label each step. Limit to 3 steps only.]
        <step 1> One sentence with the specific thinking content of this step limiting to 15 words only
        <step 2> One sentence with the specific thinking content of this step limiting to 20 words only
        <step 3> One sentence with the specific thinking content of this step limiting to 20 words only

        ### Probability Score:
        [Score from 0 to 100 of the chance of the patient developing delirium]
        """.strip()

    prompt = "\n\n".join([
        role,
        task,
        knowledge_block,
        knowledge_instructions,
        patient_block,
        patient_instructions,
        output_format,
    ])

    return prompt
\end{lstlisting}